\documentclass[runningheads]{llncs}

\usepackage{eccv}
\usepackage{eccvabbrv}

\usepackage{graphicx}
\usepackage{booktabs}

\PassOptionsToPackage{colorlinks,
                      linkcolor=eccvblue,
                      anchorcolor=red,
                      citecolor=purple,
                      urlcolor=red}{hyperref}

\usepackage[accsupp]{axessibility}  %

\usepackage{orcidlink}
\usepackage[utf8]{inputenc} %
\usepackage[T1]{fontenc}    %
\usepackage{url}            %
\usepackage{booktabs}       %
\usepackage{amsfonts}       %
\usepackage{nicefrac}       %
\usepackage{microtype}      %
\usepackage{xcolor}         %
\usepackage{amsmath}
\usepackage{tabularx}
\usepackage{graphicx} 
\usepackage{algorithm}
\usepackage{algpseudocode}
\usepackage{titletoc}
\usepackage{amssymb}
\usepackage{multirow}
\usepackage{parskip}
\usepackage{pifont}
\usepackage{array} %
\usepackage{adjustbox}

\definecolor{my_green}{RGB}{51,102,0}
\definecolor{my_red}{RGB}{204, 0, 0}

\definecolor{paired-light-blue}{RGB}{198, 219, 239}
\definecolor{paired-dark-blue}{RGB}{49, 130, 188}
\definecolor{paired-light-orange}{RGB}{251, 208, 162}
\definecolor{paired-dark-orange}{RGB}{230, 85, 12}
\definecolor{paired-light-green}{RGB}{199, 233, 193}
\definecolor{paired-dark-green}{RGB}{49, 163, 83}
\definecolor{myblue}{RGB}{218,232,252}
\definecolor{mygray}{RGB}{220,220,220}

\newcommand{\myparagraph}[1]{\textbf{#1}\hspace{1.8ex}}

\usepackage{colortbl}
\usepackage{enumitem}
\usepackage{bm}

\usepackage{listings}
\usepackage{wrapfig}
\usepackage{bbm}
\usepackage{gradient-text}
\usepackage[most,skins,theorems]{tcolorbox}
\tcbset{
  aibox/.style={
    width=\linewidth,
    top=7pt,
    bottom=2pt,
    colback=blue!4!white,
    colframe=black,
    colbacktitle=black,
    enhanced,
    center,
    attach boxed title to top left={yshift=-0.1in,xshift=0.15in},
    boxed title style={boxrule=0pt,colframe=white,},
  }
}
\newtcolorbox{AIbox}[2][]{aibox,title=#2,#1}

\newcommand{\ours}{\texorpdfstring{\gradientRGB{SpecEyes}{29,78,216}{20,184,166}}{SpecEyes}\xspace}
\begin{document}

\title{\ours: Accelerating Agentic Multimodal LLMs via Speculative Perception and Planning} 

\titlerunning{SpecEyes}

\author{Haoyu Huang\inst{1}\textsuperscript{*} \and
Jinfa Huang\inst{2}\textsuperscript{*} \and
Zhongwei Wan\inst{3} \and
Xiawu Zheng\inst{1} \and
\\
Rongrong Ji\inst{1}$^{\dagger}$ \and
Jiebo Luo\inst{2}$^{\dagger}$}

\authorrunning{H.~Huang et al.}

\institute{
Xiamen University, Xiamen, Fujian, China \\ \and
University of Rochester, Rochester, NY, USA \\ \and
The Ohio State University, Columbus, OH, USA
\email{\{huanghaoyu@stu.,rrji@\}xmu.edu.cn}, \email{\{jhuang90@ur,jluo@cs\}.rochester.edu} 
}

\renewcommand{\thefootnote}{}
\footnotetext[0]{$^*$ Equal Contributors \quad $^\dagger$ Corresponding Authors}

\maketitle

{\centering\small\textbf{Code:} \url{https://github.com/MAC-AutoML/SpecEyes}\par\smallskip}

\begin{abstract}
Agentic multimodal large language models (MLLMs) (\eg, OpenAI o3~\cite{openai2025introducing} and Gemini Agentic Vision~\cite{doshi2026agentic}) achieve remarkable reasoning capabilities through the iterative invocation of visual tools. However, the cascaded perception, reasoning, and tool-calling loops introduce significant sequential overhead. This overhead, termed agentic depth, incurs prohibitive latency and seriously limits system-level concurrency. To this end, we propose \ours, an \emph{agentic-level} speculative acceleration framework that breaks this sequential bottleneck. 
Our key insight is that a lightweight MLLM can plan a tool-free execution path to directly answer many queries, bypassing the expensive tool-use loop.
To regulate this speculative planning, we introduce a cognitive gating mechanism based on answer separability, which quantifies the model's confidence in self-verification without requiring oracle labels.
Furthermore, we design a heterogeneous parallel funnel that exploits the small model's stateless concurrency to mask the large model's stateful serial execution, thereby maximizing system throughput. 
Extensive experiments on V* Bench, HR-Bench, and POPE demonstrate that \ours{} achieves 
$1.1-3.35\times$ speedup over the baseline while preserving or even improving accuracy, thereby boosting serving throughput under concurrent workloads.
\end{abstract}

\section{Introduction}
\label{sec:intro}

Multimodal large language models (MLLMs) have undergone a paradigm shift, from static, single-pass visual perception to dynamic, \emph{agentic} interaction with the visual world.
Early MLLMs encode an image once and generate a response in a single forward pass, treating vision as a passive input channel.
Recent breakthroughs~\cite{zheng2025deepeyes,hong2025deepeyesv2,zhang2025thyme,Song2025CodeDanceAD,guo2025thinkingwithprogrammingvision,lin2026moe} fundamentally alter this design: models actively invoke external perception tools (\eg zoom-in, crop, OCR) during reasoning, forming iterative loops of perception, reasoning, and tool calling that progressively refine their understanding.
This agentic paradigm has demonstrated remarkable capabilities on challenging visual tasks that require fine-grained inspection, multi-step compositional reasoning, and active information seeking~\cite{Lai2025Minio3SU,yang2026deepreliableadvancingmultiturn,SenseNova-MARS,ma2025benchmarkingabstractreasoningabilities,ma2026a2rbenchautomaticparadigmformally,xie2026socialomni}.

\begin{figure}[t]
  \centering
  \includegraphics[width=0.8\textwidth]{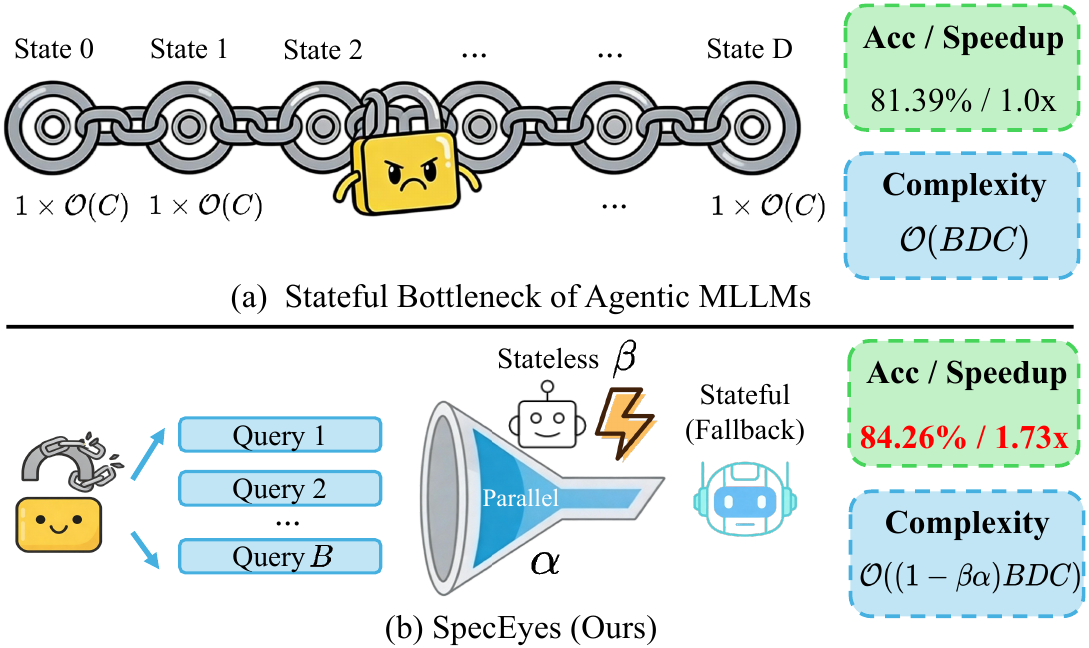} %
  \caption{\textbf{Motivation and overview of \ours.}
  \textbf{Top:} 
  Agentic MLLMs evaluate each query via a Markovian sequence of stateful tool invocations of depth $D$. This strict causal dependency prohibits parallelization, imposing a serving complexity of $\mathcal{O}(BDC)$ for $B$ queries, where $C$ denotes the tool per-step inference cost.
  \textbf{Bottom:} \ours enables agentic-level speculative bypass with a stateless small model and an answer-separability gate. Here, $\beta$ is the fraction of tool-free candidates after screening (\cref{sec:parallel}) and $\alpha$ is the acceptance rate of speculative answers among them (\cref{sec:speceyes,sec:gating}), averaging 80\% and 71\% across all benchmarks, respectively.
  All reported accuracy and speedup values are averaged across V*~\cite{vstar}, HR-Bench~\cite{hrbench}, and POPE~\cite{pope}.}
  \label{fig:fig1}
\end{figure}

However, the mechanism that empowers agentic MLLMs simultaneously introduces a severe \emph{efficiency crisis}. As shown in Fig.~\ref{fig:fig1}, each query triggers a cascade of tool-calling steps, a quantity we term the \emph{agentic depth} $D$, in which each step depends on the observation from the previous step.
This strict data dependency inflicts a \textbf{dual disaster} on system performance:
\textbf{(i)~Latency explosion}: the end-to-end response time for a single query grows linearly with $D$, since each reasoning-and-tool cycle must complete before the next can begin;
\textbf{(ii)~Concurrency collapse}: because each query's tool-use chain mutates a per-query state, batching efficiency is severely bottlenecked, the agentic model can only advance one step at a time per query, leaving massive hardware parallelism idle.
Therefore, these effects render agentic MLLMs orders of magnitude slower than non-agentic counterparts, posing a fundamental barrier to real-world deployment.

Existing approaches to efficient reasoning fall short of addressing this bottleneck.
Token-level speculative decoding~\cite{pan2025specreason,Huang2026RelayLLMER} accelerates individual generation steps by letting a small draft model propose tokens for a larger model to verify. 
However, these methods still operate \emph{within} a fixed reasoning trajectory: the \emph{agentic pipeline itself}, \ie, the multi-turn loop of perception and reasoning, remains fully serial and every tool must still be invoked in sequence. 
Moreover, the additional draft/verification interaction often expands the generated traces (longer token sequences and extra turns), introducing non-trivial overhead that can offset the per-step speedup in practice.
Similarly, multimodal token pruning~\cite{endo2025feather,li2025herorethinkingvisualtoken,he2024zipvl,wang2025fouriervlm} and temporal compression~\cite{fu2025framefusion,Hu2025ThinkingWD} reduce per-step compute within a fixed model, yet they do not eliminate the repeated tool invocations that dominate agentic latency.
In short, all prior methods operate \emph{within} the agentic loop, none question whether the loop itself is necessary for every query.

In this paper, we make a conceptual leap: we lift the speculative paradigm from the token/semantic level to the \textbf{agentic level}.
Our key observation is that a large fraction of queries directed at agentic MLLMs do \emph{not} actually require deep tool-assisted reasoning. Instead, a lightweight, tool-free vision model can answer them correctly using only the original image, provided we can reliably identify which queries fall into this category.
This motivates a heterogeneous ``\emph{think fast, think slow}'' architecture: a small non-agentic model rapidly generates speculative answers via intuition (fast thinking), while the large agentic model is reserved for queries that genuinely require multi-step tool interaction (slow thinking).

We instantiate this idea by introducing \ours, an \emph{agentic-level} speculative acceleration framework for multimodal reasoning.
It comprises three tightly integrated components:
\textbf{(1)~A four-phase speculative pipeline} (\cref{sec:speceyes}) that routes each query through heuristic tool-use judgment, small-model speculation, confidence-based switching, and agentic fallback.
\textbf{(2)~Cognitive gating} (\cref{sec:gating}) via a novel \emph{answer separability} metric $S_{\text{sep}}$ that measures the competitive margin among top-$K$ logits, providing a calibration-free, scale-invariant decision boundary for trusting the small model's output.
\textbf{(3)~A heterogeneous parallel serving architecture} (\cref{sec:parallel}) that runs the stateless small model concurrently and forwards only low-confidence queries to the agentic model, converting the speculative acceptance rate into multiplicative throughput gains.
Extensive experiments on V*~Bench, HR-Bench, and POPE show that \ours preserves the full accuracy of the agentic pipeline while substantially reducing latency and improving throughput. Overall, the main contributions are as follows:

\begin{itemize}
\item We identify and formalize the \emph{stateful bottleneck} of agentic MLLMs, showing that data dependency inherent in tool-use chains imposes a fundamental barrier to both per-query latency and system-level concurrency.
\item We propose \ours, the first framework that lifts speculative acceleration from the token level to the \emph{agentic level}, bypassing entire tool-use loop for queries that do not require it while preserving full accuracy.
\item We introduce \emph{cognitive gating} based on answer separability among top-$K$ logits, providing a label-free, scale-invariant criterion for the small model to decide when to trust its own output versus escalating to the agentic model.
\item We design a \emph{heterogeneous parallel funnel} that exploits the stateless nature of the small model to achieve concurrent query processing, yielding multiplicative throughput improvements proportional to the speculative acceptance rate.
\end{itemize}

\section{Related Work}
\label{sec:related_work}

\myparagraph{Agentic Multimodal Large Language Models.}
Agentic reasoning in language models originates from tool-augmented frameworks that interleave action generation with external feedback~\cite{yao2022react,schick2023toolformer,shen2023hugginggpt,yu2025recode}.
Building on this, multimodal large language models (MLLMs) have adopted a similar agentic paradigm, enabling active interleaving of perception and reasoning through external visual tools rather than relying on passive single-pass encoding.
Early large-scale MLLMs~\cite{li2023blip,alayrac2022flamingo,dai2023instructblip,bai2023qwen,team2023gemini} established the backbone architectures upon which agentic extensions are built.
DeepEyes~\cite{zheng2025deepeyes} demonstrates that reinforcement learning can train models to call perception tools during reasoning; subsequent work enables executable reasoning via code generation and visual manipulation~\cite{zhang2025thyme,Song2025CodeDanceAD,hong2025deepeyesv2,guo2025thinkingwithprogrammingvision,zhang2025skywork,zhao2026pyvision,team2026kimi,hou2026codevcodeimagesfaithful, xie2025training}, and further scales agentic depth through multi-turn interaction, self-reflection, and reinforcement-learning-based agent optimization~\cite{Lai2025Minio3SU,yang2026deepreliableadvancingmultiturn,SenseNova-MARS,peng2025skyworkr1v,lian2025ui}.
Despite their effectiveness, these methods rely on deeply sequential perception–reasoning tool loops, incurring substantial latency and limited concurrency, a system-level bottleneck that prior work largely overlooks.

\myparagraph{Efficient Reasoning.}
Token-level speculative decoding~\cite{leviathan2023fast,cai2024medusa,chen2023accelerating,xia-etal-2023-speculative,li2024eagle1,li2024eagle2,li2025eagle3, zhang2024draft, xia2024swift, yang2025longspec, xu2025specee} accelerates generation by having a small draft model propose tokens for a larger model to verify.
Recent extensions apply this idea to collaborative reasoning: SpecReason~\cite{pan2025specreason} delegates simpler steps to a lightweight model verified via semantic consistency; RelayLLM~\cite{Huang2026RelayLLMER} dynamically invokes a stronger expert at critical steps; ATTS~\cite{xiong2026attsasynchronoustesttimescaling} further explores asynchronous test-time scaling by dynamically allocating computation under uncertainty;
DSP~\cite{guan2025dynamic} speculatively drafts and verifies agent actions \emph{within} text-only LLM-agent trajectories via online reinforcement learning,
and SpecTemp~\cite{Hu2025ThinkingWD} and Lin et al.~\cite{lin2025speculative, lin2025accelerating} reduce redundant visual processing in multimodal and interactive settings.
Adaptive computation and early-exit methods~\cite{teerapittayanon2016branchynet,kumar2025helios,chen2023ee,fan2024not,zhu2024hierarchical,luo2026video} further bypass layers for easier inputs. In contrast to methods that only optimize steps \emph{within} a fixed trajectory, our \ours speculatively bypasses the agentic loop entirely, breaking sequential bottlenecks to unlock parallel execution.

\myparagraph{Efficient Multimodal Perception.}
A parallel line of work reduces the per-step computational burden of
multimodal perception.
Frequency-based compression truncates high-frequency visual
signals~\cite{wang2025fouriervlm};
token pruning retains visually salient tokens via attention scores or
multimodal relevance~\cite{endo2025feather,li2025herorethinkingvisualtoken,
xing2024pyramiddrop,yang2025visionzip,twigvlm_shao};
and dynamic sparsification optimizes retention across
layers~\cite{he2024zipvl}.
Token merging~\cite{bolya2022token,kim2024token,wang2025efficient} reduces
sequence length by combining redundant representations, and temporal
redundancy across frames is exploited to merge or prune spatial tokens in
video settings~\cite{fu2025framefusion,chen2026waveletbasedframeselectiondetecting}.
KV-cache compression~\cite{wan2024look,wan2025meda,liu2024efficient}
additionally reduces memory and decoding cost by evicting cached visual
keys and values.
Despite these gains, all such methods operate within a monolithic model and
leave the sequential agentic pipeline intact, as the large model must still execute the full perception--reasoning loop.
In contrast, \ours targets efficiency at the \emph{agentic level}: rather than accelerating individual operations, it speculatively bypasses entire tool-use loops via a cognitively gated lightweight model, breaking sequential bottlenecks to enable high-throughput parallel execution.

\section{Methodology}
\label{sec:method}

\begin{figure}[tbp]
    \centering 
    \includegraphics[width=\textwidth]{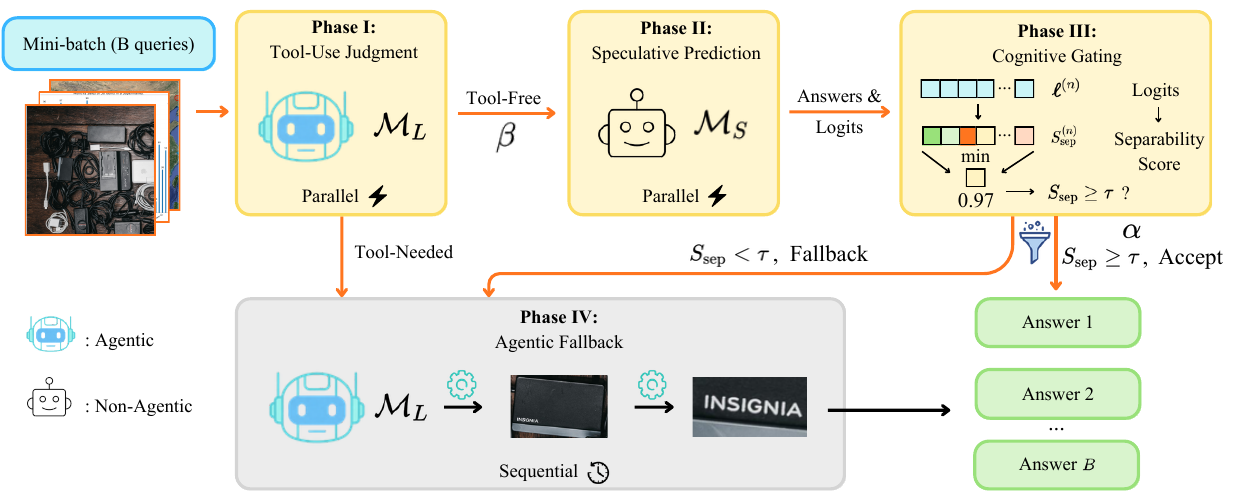}  
    \caption{\textbf{Pipeline overview of \ours.}
    A batch of $B$ queries passes through a four-phase funnel.
    \textbf{I:} $\mathcal{M}_L$ screens tool necessity, splitting queries into tool-free ($g{=}0$) and tool-required ($g{=}1$).
    \textbf{II:} A stateless $\mathcal{M}_S$ speculatively answers all tool-free queries with token-level logits.
    \textbf{III:} An answer separability score $S_{\text{sep}}$ gates each answer; those above $\tau$ are accepted directly.
    \textbf{IV:} Remaining queries fall back to the full agentic loop.
    The funnel yields $\approx\!1/(1{-}\beta\alpha)\times$ throughput speedup.}
    \label{fig:fig2}
\end{figure}

We begin by formalizing the stateful bottleneck inherent in agentic multimodal reasoning (\cref{sec:bottleneck}), then introduce SpecEyes, our four-phase speculative acceleration framework (\cref{sec:speceyes}). We detail the cognitive gating mechanism that governs speculative bypass (\cref{sec:gating}), and finally describe the heterogeneous parallel architecture that maximizes system throughput (\cref{sec:parallel}).

\subsection{Modeling the Stateful Bottleneck of Agentic MLLMs}
\label{sec:bottleneck}

\myparagraph{Preliminaries.}
We formalize an agentic multimodal large language model (MLLM) as a stateful reasoning system $\mathcal{A} = (\mathcal{S}, \mathcal{T}, \pi)$, where $\mathcal{S}$ denotes the state space, $\mathcal{T} = \{t_1, \ldots, t_N\}$ is a set of perception tools (\eg \texttt{Zoom-in}, \texttt{Crop}), and $\pi$ is policy that jointly selects tool invocations and generates reasoning tokens.

Given a query $q$ and an input image $I$, the model maintains a state trajectory $\{s_0, s_1, \ldots, s_D\}$ over $D$ reasoning steps. The initial state is $s_0 = (q, I)$. At each step $d$, the policy produces an action $a_d = \pi(s_d)$ that either invokes a tool $t \in \mathcal{T}$ or emits a final answer. When a tool is invoked, the state transitions as:
\begin{equation}
\label{eq:state_transition}
s_{d+1} = f(s_d, t_d(s_d)),
\end{equation}
where $t_d(s_d)$ applies the selected tool $t_d$ to the current visual context (\eg cropping a region of interest from $I$) and $f$ fuses the resulting observation into the next state. We refer to $D$ as the \emph{agentic depth} of the query.

\myparagraph{State Dependency and Sequential Bottleneck.}
A critical property of \cref{eq:state_transition} is that subsequent tool selections depend causally on prior observations. Concretely, let $t_{d+1} \sim \pi(\cdot \mid s_{d+1})$ be tool chosen at step $d{+}1$. Since $s_{d+1}$ contains the output of $t_d$,  Markov chain $(s_0, a_0, s_1, a_1, \ldots)$ forms a strict data dependency:
\begin{equation}
\label{eq:dependency}
p(a_{d+1} \mid s_0, a_0, \ldots, s_d) = p(a_{d+1} \mid s_d, t_d(s_d)) \neq p(a_{d+1} \mid s_0).
\end{equation}
This dependency renders the agentic pipeline inherently \emph{sequential}: step $d{+}1$ cannot begin until step $d$ completes. Consequently, the end-to-end latency for a single query scales linearly with agentic depth:
\begin{equation}
\label{eq:latency}
L_{\text{agent}}(q) = \sum_{d=0}^{D(q)} \big(\underbrace{c_{\text{llm}}}_{\text{reasoning}} + \underbrace{c_{\text{tool}}(t_d)}_{\text{perception}}\big),
\end{equation}
where $c_{\text{llm}}$ and $c_{\text{tool}}(t_d)$ denote the latency of LLM inference and tool execution at step $d$, respectively.

\myparagraph{Throughput Implication.}
At the system level, this strict serialization limits concurrency even under continuous batching (\eg, vLLM~\cite{kwon2023efficient}). While LLM forward passes for different queries at different agentic steps can share a GPU batch, the stateful tool-use loop for each query still proceeds sequentially: step $d{+}1$ of query $i$ cannot begin until step $d$ completes. Consequently, within a batch of $B$ queries, the batch-level wall-clock time is dominated by the \emph{slowest} (deepest) trajectory, as queries with heavy-tailed agentic depth stall the entire batch. The effective throughput under batched serving is:
\begin{equation}
\label{eq:throughput_bound}
\Theta_{\text{agent}}^{\text{batched}} \approx \frac{B}{\max_{i \in [B]} L_{\text{agent}}(q_i) + c_{\text{sched}}},
\end{equation}
where $c_{\text{sched}}$ is a small scheduling overhead. This bound tightens as agentic depth variance grows, since a single long-tail query forces the entire batch to wait. Speculatively converting a fraction $\beta\alpha$ of queries into stateless single-pass inferences directly shrinks the effective $\max_{i}L_{\text{agent}}(q_i)$, motivating our approach.

\subsection{SpecEyes: Agentic-Level Speculative Reasoning}
\label{sec:speceyes}

Our key insight is that not all queries require deep agentic reasoning. For a substantial fraction of inputs, a small \emph{non-agentic} MLLM, denoted $\mathcal{M}_S$, can produce a correct answer \emph{without any tool invocation}, directly from the original image $I$. SpecEyes exploits this observation through a four-phase pipeline (\cref{fig:fig2}) that speculatively bypasses expensive tool chains whenever $\mathcal{M}_S$ is sufficiently confident, and falls back to the full agentic model $\mathcal{M}_L$ otherwise.

We denote the small non-agentic model as $\mathcal{M}_S$ and the large agentic MLLM as $\mathcal{M}_L = \mathcal{A}$. The four phases are detailed below.

\myparagraph{Phase~I: Heuristic Tool-Use Judgment.}
Given a query $q$ and image $I$, the large agentic model $\mathcal{M}_L$ first determines whether tool invocation is necessary. We prompt $\mathcal{M}_L$ with a lightweight binary classification head:
\begin{equation}
\label{eq:tool_judgment}
g(q, I) = \mathcal{M}_L\!\left(q, I;\; \mathcal{P}_{\text{judge}}\right) \in \{0, 1\},
\end{equation}
where $\mathcal{P}_{\text{judge}}$ is a prompt instructing the model to assess
tool necessity, $g = 0$ indicates that $\mathcal{M}_L$ judges the query to be answerable from the global image alone, and $g = 1$ indicates a potential need for tool-assisted perception. Queries with $g = 0$ proceed directly to Phase~II; queries with $g = 1$ are immediately forwarded to Phase~IV (agentic fallback). Although Phase~I is executed by $\mathcal{M}_L$, it generates only a single binary token with no tool invocation, incurring negligible overhead. We use $\mathcal{M}_L$ rather than $\mathcal{M}_S$ because its tool-calling capability makes it a more reliable judge of tool necessity, yielding more accurate screening.

\myparagraph{Phase~II: Speculative Prediction.}
For queries passing Phase~I (\ie, $g = 0$), $\mathcal{M}_S$ directly generates an answer $\hat{y}_S$ along with the full output logit distribution:
\begin{equation}
\label{eq:spec_pred}
\hat{y}_S, \;\{\boldsymbol{\ell}^{(n)}\}_{n=1}^{|\hat{y}_S|} = \mathcal{M}_S(q, I),
\end{equation}
where $\boldsymbol{\ell}^{(n)} \in \mathbb{R}^{|\mathcal{V}|}$ is the logit vector over the vocabulary $\mathcal{V}$ for the $n$\textsuperscript{th} generated token. Crucially, this inference is \emph{stateless}: it requires no tool execution and can be performed concurrently for all queries in the batch.

\myparagraph{Phase~III: Small MLLM Confidence Switching.}
The logits from Phase~II are passed to a \emph{cognitive gating} function $S_{\text{sep}}$ (detailed in \cref{sec:gating}) that quantifies the answer confidence of $\mathcal{M}_S$ without requiring ground-truth labels. We compute a scalar separability score for the speculative answer $\hat{y}_S$:
\begin{equation}
\label{eq:switching}
\text{decision} =
\begin{cases}
\texttt{accept}\ \hat{y}_S, & \text{if } S_{\text{sep}}(\hat{y}_S) \geq \tau, \\[4pt]
\texttt{fallback to } \mathcal{M}_L, & \text{if } S_{\text{sep}}(\hat{y}_S) < \tau,
\end{cases}
\end{equation}
where $\tau$ is a threshold selected from a coarse operating-point grid. Accepted answers are returned immediately, completely bypassing the agentic pipeline; rejected queries proceed to Phase~IV.

\myparagraph{Phase~IV: Agentic Fallback.}
Queries that fail confidence switching are routed to the full agentic model $\mathcal{M}_L$, which executes the complete stateful perception-reasoning loop:
\begin{equation}
\label{eq:fallback}
\hat{y}_L = \mathcal{M}_L(q, I) = \pi\big(s_0 \xrightarrow{t_0} s_1 \xrightarrow{t_1} \cdots \xrightarrow{t_{D-1}} s_D\big).
\end{equation}
The agentic model retains full access to all tools $\mathcal{T}$ and performs multi-step reasoning at the cost of sequential latency $L_{\text{agent}}(q)$. By design, Phase~IV serves as a \emph{safety net}: routing low-confidence queries back to the full agentic pipeline substantially mitigates potential accuracy loss, even if a marginal gap may remain due to the imperfect nature of the gating mechanism.

\myparagraph{End-to-End Latency.}
Let $\beta \in [0,1]$ denote the tool-free screening ratio from Phase~I
and $\alpha \in [0,1]$ the cognitive gate acceptance rate from Phase~III.
All queries incur the judgment cost $c_J$, only the $\beta$ fraction passing
Phase~I additionally incurs the small model cost $c_S$, the remaining
$(1-\beta\alpha)$ fraction forwarded to $M_L$ pays the full agentic cost
$L_{\text{agent}}$. Therefore, the expected per-query latency is:
\begin{equation}
    \mathbb{E}\!\left[L_{\text{SpecEyes}}\right]
    = c_J + \beta\, c_S + \bigl(1-\beta\alpha\bigr)\, L_{\text{agent}},
    \label{eq:e2elatency}
\end{equation}
where $c_J + \beta c_S \ll L_{\text{agent}}$.
When $\beta\alpha$ is large (\eg $\beta\alpha > 0.6$),
the expected latency is dominated by the lightweight front-end cost,
yielding substantial speedups over the purely agentic baseline.

\subsection{Small MLLM Cognitive Gating via Answer Separability}
\label{sec:gating}

The effectiveness of SpecEyes hinges critically on the quality of the confidence switching mechanism in Phase~III. We now introduce the \emph{answer separability} score $S_{\text{sep}}$ that serves as the cognitive gate.

\myparagraph{Limitations of Probability-Based Confidence.}
A common probability-based confidence for sequence generation aggregates per-token max-softmax probabilities via the geometric mean\cite{zhao2025stitch}. Concretely, for the $n$-th generated token with logits $\boldsymbol{\ell}^{(n)}$, we define the maximum softmax probability $p_{\max}^{(n)}$ as:
\begin{equation}
    p_{\max}^{(n)} = \max_{v \in \mathcal{V}} \sigma(\boldsymbol{\ell}^{(n)})_v,
\end{equation}
where $\sigma(\cdot)$ denotes the softmax operator and $\mathcal{V}$ is the vocabulary.
The overall confidence is computed as:
\begin{equation}
\label{eq:prob_conf}
S_{\text{log}}(\hat{y}_S)
= \exp\!\left(\frac{1}{|\hat{y}_S|}\sum_{n=1}^{|\hat{y}_S|}\log p_{\max}^{(n)}\right),
\end{equation}
which corresponds to the geometric mean of $\{p_{\max}^{(n)}\}$.
However, $S_{\text{log}}$ remains unreliable for gating: (1)~it inherits the well-known miscalibration of softmax, where large logit magnitudes can yield overconfident probabilities; (2)~token-wise $p_{\max}^{(n)}$ can be spuriously high for low-entropy or nearly-deterministic positions (e.g., punctuation, formatting tokens), and the geometric aggregation does not explicitly measure how well the top prediction is separated from strong competitors. These issues increase the risk of false acceptance in our speculative bypass.

\myparagraph{Answer Separability Score.}
Instead of relying on the raw softmax probability, we design a metric that measures the \emph{decision margin} between the top prediction and its competitors. For the $n$\textsuperscript{th} generated token with logit vector $\boldsymbol{\ell}^{(n)}$, let $\ell_{[1]}^{(n)} \geq \ell_{[2]}^{(n)} \geq \cdots \geq \ell_{[|\mathcal{V}|]}^{(n)}$ be the sorted logits in descending order. We define the \emph{token-level separability} as:
\begin{equation}
\label{eq:token_sep}
S_{\text{sep}}^{(n)} = \frac{\ell_{[1]}^{(n)} - \mu_K^{(n)}}{\sigma_K^{(n)} + \epsilon},
\end{equation}
where $\mu_K^{(n)}$ and $\sigma_K^{(n)}$ are the mean and standard deviation of the top-$K$ logits $\{\ell_{[1]}^{(n)}, \ldots, \ell_{[K]}^{(n)}\}$, and $\epsilon > 0$ is a small constant for numerical stability. Intuitively, $S_{\text{sep}}^{(n)}$ quantifies how far the leading logit stands apart from its nearest competitors: a large value indicates a clear decision boundary, while a small value signals ambiguity among top candidates.

Compared to softmax probability, $S_{\text{sep}}^{(n)}$ offers two key advantages: (i)~it is \emph{scale-invariant}, since both the numerator and denominator scale linearly with logit magnitude, neutralizing the calibration artifacts of softmax; (ii)~it explicitly models the \emph{competitive landscape} among top candidates via the variance term $\sigma_K^{(n)}$, providing a more informative confidence signal.

\myparagraph{Token-to-Answer Aggregation.}
The token-level score $S_{\text{sep}}^{(n)}$ must be aggregated across all $|\hat{y}_S|$ generated tokens to obtain an answer-level confidence. We consider three natural aggregation strategies:
\begin{equation}
\label{eq:aggregation}
S_{\text{sep}}^{\text{mean}} = \frac{1}{|\hat{y}_S|}\sum_{n=1}^{|\hat{y}_S|} S_{\text{sep}}^{(n)}, \quad
S_{\text{sep}}^{\text{min}} = \min_{n \in [|\hat{y}_S|]} S_{\text{sep}}^{(n)}, \quad
S_{\text{sep}}^{\text{bottom-}r} = \frac{1}{|\mathcal{B}|}\sum_{n \in \mathcal{B}} S_{\text{sep}}^{(n)},
\end{equation}
where $\mathcal{B}$ is the index set of the bottom-$r$ fraction of tokens with the smallest $S_{\text{sep}}^{(n)}$ values, \ie, $|\mathcal{B}|=\lceil r\,|\hat{y}_S|\rceil$ for a ratio $r \in (0,1)$ chosen empirically. The aggregated score is then normalized via a sigmoid function. We adopt $\min$ aggregation as the default strategy, based on the following risk-theoretic argument.
\begin{proposition}
\label{prop:min}
Let $\hat{y}_S = (y_1, \ldots, y_{|\hat{y}_S|})$ be the speculative answer. Define the answer-level error event $\mathcal{E} = \bigcup_{n} \mathcal{E}_n$, where $\mathcal{E}_n$ denotes the event that token $y_n$ is incorrect. Then:
\begin{equation}
\label{eq:min_bound}
P(\mathcal{E}) = P\!\left(\bigcup_{n} \mathcal{E}_n\right) \leq \sum_{n} P(\mathcal{E}_n).
\end{equation}
\end{proposition}

Under the assumption that each $P(\mathcal{E}_n)$ is monotonically decreasing in $S_{\emph{sep}}^{(n)}$, a plausible working assumption, thresholding on $\min_n S_{\emph{sep}}^{(n)}$ ensures every token exceeds the confidence threshold, providing the tightest bound on $P(\mathcal{E})$. We therefore treat $S_{\text{sep}}^{\text{min}}$ as an \emph{empirically motivated} default: it consistently achieves the highest matched-speed accuracy among all four aggregation variants across every benchmark . Intuitively, the $\min$ strategy acts as a \emph{worst-case guard}: it triggers fallback whenever \emph{any} token in the answer exhibits low separability. This is conservative by design, prioritizing precision (\ie, avoiding false acceptances) to preserve the accuracy guarantee of the agentic pipeline. 

\subsection{Heterogeneous Parallelism for Throughput Acceleration}
\label{sec:parallel}

Beyond per-query latency reduction, \ours enables system-level throughput gains by organizing the four phases into a heterogeneous parallel funnel.

\myparagraph{Batch-Parallel Front-End.}
We serve requests in batches of size $B$. Let $\beta\in[0,1]$ be the fraction of queries that Phase~I screens as tool-free ($g{=}0$) and $\alpha\in[0,1]$ be the acceptance rate of the cognitive gate among those candidates. Both screening (Phase~I, latency $c_J$) and speculative inference (Phase~II, latency $c_S$) are stateless single-turn forward passes and therefore fully batch-parallelizable, giving a parallel front-end cost of $c_J + c_S$.

\myparagraph{Funnel-Shaped Serving.}
Accepted queries ($\alpha, \beta, B$) are returned immediately; the remaining \emph{residual set} $\mathcal{R}$, consisting of gating-rejected and tool-required queries, falls back to sequential agentic execution:
\begin{equation}
\label{eq:funnel}
\begin{aligned}
&\underbrace{B}_{\text{batch}}
\xrightarrow{\;\mathcal{M}_L~\text{screen (par.)}\;}
\underbrace{\beta B}_{g=0}
\;+\;
\underbrace{(1-\beta)B}_{g=1}
\\
&\underbrace{\beta B}_{g=0}
\xrightarrow{\;\mathcal{M}_S~\text{speculate (par.)}\;}
\underbrace{\alpha\beta B}_{\text{accept}}
\;+\;
\underbrace{(1-\alpha)\beta B}_{\text{reject}}
\\
&\underbrace{(1-\beta)B+(1-\alpha)\beta B}_{\mathcal{R}}
\xrightarrow{\;\mathcal{M}_L~\text{agentic (seq.)}\;}
\underbrace{(1-\beta\alpha)B}_{\text{fallback}} .
\end{aligned}
\end{equation}
Under continuous batching (\eg, vLLM~\cite{kwon2023efficient}), throughput scales approximately linearly with the number of queries served through the agentic path. SpecEyes converts a $\beta\alpha$ fraction of queries into stateless single-pass inferences that bypass the agentic loop entirely, reducing the effective agentic residual from $B$ to $(1{-}\beta\alpha)B$. Since $c_J + c_S \ll \bar{L}_{\text{agent}}$, the per-batch cost is dominated by the agentic fallback on $|\mathcal{R}|=(1{-}\beta\alpha)B$ residual queries, yielding a throughput speedup is:
\begin{equation}
\label{eq:speedup}
\Theta_{\text{SpecEyes}}\,/\,\Theta_{\text{agent}}
\;\approx\; {1}/({1-\beta\alpha}),
\end{equation}
jointly governed by the screening ratio $\beta$ and the gate acceptance rate $\alpha$. Importantly, continuous batching multiplies throughput proportionally for \emph{both} the baseline and \ours, so it does not alter the speedup ratio.

\section{Experiment}
\label{sec:exp}

\subsection{Experiment Setups}
\label{sec:setup}

\myparagraph{Benchmarks and Baselines.}
We evaluate \ours on three multimodal benchmarks spanning fine-grained perception, high-resolution understanding, and hallucination robustness.
V*~\cite{vstar} provides two multiple-choice subsets: Direct Attributes (115 questions) for attribute recognition and Relative Position (76 questions) for spatial reasoning.
HR-Bench~\cite{hrbench} tests high-resolution perception with 4K and 8K subsets (800 questions each).
POPE~\cite{pope} is a yes/no hallucination probe with Adversarial, Popular, and Random splits (3\,000 questions each).
All benchmarks are evaluated by accuracy.
The small non-agentic model $M_S$ is Qwen3-VL-2B~\cite{qwen3technicalreport}, the large agentic model $M_L$ is instantiated with DeepEyes~\cite{zheng2025deepeyes} and Thyme~\cite{zhang2025thyme}, both capped at 5 tool-use steps per query.

\myparagraph{Implementation Details.}
All models use greedy decoding (temperature 0), and all reported latencies include tool execution time.
For cognitive gating (\cref{sec:gating}), we set $K{=}64$, $\epsilon{=}10^{-6}$, and adopt min-token aggregation; for the bottom aggregation variant, we set the bottom fraction to $r{=}0.2$, 
inspired by \cite{fu2025deep}.
The gating threshold $\tau$ is selected as follows: we run $\mathcal{M}_S$ once on a random 10\% sample of each benchmark solely to \emph{visualize} the empirical $S_\text{sep}$ distribution and define a coarse search range. We then evenly sample a fixed grid of operating points from this range, the reported $\tau$ for each variant is chosen from this grid with \emph{no per-sample or per-benchmark optimization}, so any overlap with the test set is immaterial and the results do not constitute an upper bound.
All experiments run on a single NVIDIA A100 40GB GPU.

\definecolor{oursrow}{RGB}{226, 243, 227}
\definecolor{variantrow}{RGB}{240, 248, 255}
\definecolor{groupbg}{RGB}{235, 235, 235}
\definecolor{speedgreen}{RGB}{0, 120, 0}
\definecolor{speedred}{RGB}{190, 40, 40}

\newcommand{\best}[1]{\textbf{#1}}
\newcommand{\second}[1]{\underline{#1}}
\newcommand{\spd}[1]{\textcolor{speedgreen}{#1$\times$}}
\newcommand{\slo}[1]{\textcolor{speedred}{#1$\times$}}
\newcommand{\bas}[1]{#1$\times$}
\newcommand{\na}{\textcolor{gray}{--}}

\begin{table*}[t]
\centering
\caption{
\textbf{Main results on V*, HR-Bench, and POPE datasets.}
Spd.\ = wall-clock speedup over each base model (\textcolor{speedgreen}{green}: faster; \textcolor{speedred}{red}: slower). Bold indicates the best accuracy within each group, and highlighted rows represent our recommended variants.
SpecEyes (min) provides the best speedup-accuracy trade-off across both agentic backbones. \textit{Backbone} (w/o tools) uses the same large model with tools disabled.
}
\label{tab:main_results}
\setlength{\tabcolsep}{3.6pt}
\renewcommand{\arraystretch}{1.15}
\resizebox{\textwidth}{!}{%
\begin{tabular}{@{} l cc cc cc cc cc cc cc cc @{}}
\toprule
\multirow{3}{*}{\textbf{Method}} 
& \multicolumn{4}{c}{\textbf{V*}} 
& \multicolumn{4}{c}{\textbf{HR-Bench}} 
& \multicolumn{6}{c}{\textbf{POPE}} 
& \multicolumn{2}{c}{\multirow{2}{*}{\textbf{Avg.}}} \\
\cmidrule(lr){2-5} \cmidrule(lr){6-9} \cmidrule(lr){10-15}
& \multicolumn{2}{c}{Attr.} 
& \multicolumn{2}{c}{Pos.} 
& \multicolumn{2}{c}{4K} 
& \multicolumn{2}{c}{8K} 
& \multicolumn{2}{c}{Adv.} 
& \multicolumn{2}{c}{Pop.} 
& \multicolumn{2}{c}{Rand.} 
& \multicolumn{2}{c}{} \\
\cmidrule(lr){2-3} \cmidrule(lr){4-5} 
\cmidrule(lr){6-7} \cmidrule(lr){8-9} 
\cmidrule(lr){10-11} \cmidrule(lr){12-13} 
\cmidrule(lr){14-15} \cmidrule(lr){16-17}
& Acc. & Spd. & Acc. & Spd. 
& Acc. & Spd. & Acc. & Spd. 
& Acc. & Spd. & Acc. & Spd. 
& Acc. & Spd. 
& Acc. & Spd. \\
\midrule

Qwen3-VL-2B (draft only)
& 77.39 & \spd{5.44} 
& 82.89 & \spd{5.31} 
& 71.38 & \spd{3.20} 
& 68.00 & \spd{2.90} 
& 82.56 & \spd{4.20} 
& 83.80 & \spd{3.78} 
& 86.47 & \spd{4.07} 
& 78.93 & \spd{4.13} \\

\midrule
\multicolumn{17}{@{}l}{{\strut\textbf{\textit{Based on DeepEyes~\cite{zheng2025deepeyes}}}}} \\

\rowcolor{groupbg}
DeepEyes (w tools)
& 90.43 & \bas{1.00} 
& 82.89 & \bas{1.00} 
& 75.85 & \bas{1.00} 
& 71.43 & \bas{1.00} 
& 78.43 & \bas{1.00} 
& 81.90 & \bas{1.00} 
& 88.83 & \bas{1.00} 
& 81.39 & \bas{1.00} \\

DeepEyes (w/o tools)
& 80.87 & \spd{4.08}
& 73.68 & \spd{4.18}
& 75.25 & \spd{2.71}
& 72.00 & \spd{2.53}
& 46.90 & \spd{3.78}
& 49.33 & \spd{3.60}
& 48.20 & \spd{3.81}
& 63.75 & \spd{3.53} \\

SpecReason~\cite{pan2025specreason} 
& 80.19 & \slo{0.61} 
& 73.91 & \slo{0.38} 
& 80.43 & \slo{0.44} 
& 72.54 & \slo{0.42} 
& 49.10 & \slo{0.38} 
& 51.55 & \slo{0.38} 
& 60.20 & \slo{0.37} 
& 66.85 & \slo{0.43} \\

\cmidrule(l){1-17}

\rowcolor{variantrow}
\quad {$\triangleright$}~SpecEyes (log)
& \second{83.48} & \spd{2.06} 
& \second{88.16} & \spd{2.05} 
& 73.71 & \spd{1.35} 
& 69.67 & \spd{1.28} 
& 83.97 & \spd{1.89} 
& 86.70 & \spd{1.95} 
& \best{90.50} & \spd{2.05} 
& 82.31 & \spd{1.80} \\

\rowcolor{variantrow}
\quad {$\triangleright$}~SpecEyes (mean) 
& 78.26 & \spd{2.89} 
& 84.21 & \spd{3.35} 
& 71.62 & \spd{1.88} 
& 67.38 & \spd{1.77} 
& \best{85.13} & \spd{2.06} 
& \best{87.00} & \spd{2.10} 
& \second{90.13} & \spd{2.14} 
& 80.53 & \spd{2.31} \\

\rowcolor{variantrow}
\quad {$\triangleright$}~SpecEyes (bottom-$r$) 
& \second{83.48} & \spd{2.13} 
& 84.21 & \spd{2.12} 
& \second{75.22} & \spd{1.20} 
& \second{71.18} & \spd{1.04} 
& \best{85.13} & \spd{2.08} 
& \best{87.00} & \spd{2.08} 
& \second{90.13} & \spd{2.11} 
& \second{82.34} & \spd{1.82} \\

\rowcolor{oursrow}
\quad {$\triangleright$}~\textbf{SpecEyes (min)} 
& \best{90.43} & \spd{1.53} 
& \best{89.47} & \spd{1.90} 
& \best{75.85} & \spd{1.13} 
& \best{71.80} & \spd{1.08} 
& \best{85.13} & \spd{2.13} 
& \best{87.00} & \spd{2.15} 
& \second{90.13} & \spd{2.19} 
& \best{84.26} & \spd{1.73} \\

\midrule
\multicolumn{17}{@{}l}{{\strut\textbf{\textit{Based on Thyme~\cite{zhang2025thyme}}}}} \\

\rowcolor{groupbg}
Thyme (w tools)
& 86.96 & \bas{1.00} 
& 82.89 & \bas{1.00} 
& 77.72 & \bas{1.00} 
& 72.43 & \bas{1.00} 
& 81.32 & \bas{1.00} 
& 84.53 & \bas{1.00} 
& 90.17 & \bas{1.00} 
& 82.29 & \bas{1.00} \\

Thyme (w/o tools)
& 84.35 & \spd{2.81}
& 76.32 & \spd{2.56}
& 74.25 & \spd{1.85}
& 69.88 & \spd{1.97}
& 77.77 & \spd{3.51}
& 78.17 & \spd{3.32}
& 79.93 & \spd{2.99}
& 77.24 & \spd{2.72} \\

SpecReason~\cite{pan2025specreason} 
& 89.57 & \slo{0.48} 
& 75.00 & \slo{0.53} 
& 80.01 & \slo{0.52} 
& 81.02 & \slo{0.51} 
& 84.62 & \slo{0.46} 
& 85.97 & \slo{0.43} 
& 90.27 & \slo{0.46} 
& 83.78 & \slo{0.48} \\

\cmidrule(l){1-17}

\rowcolor{variantrow}
\quad {$\triangleright$}~SpecEyes (log)
& \second{80.87} & \spd{1.82} 
& \best{82.89} & \spd{1.45} 
& 74.97 & \spd{1.13} 
& 70.84 & \spd{1.06} 
& 85.76 & \spd{1.68} 
& 87.80 & \spd{1.67} 
& \best{91.47} & \spd{1.59} 
& \second{82.09} & \spd{1.49} \\

\rowcolor{variantrow}
\quad {$\triangleright$}~SpecEyes (mean) 
& 77.39 & \spd{2.34} 
& 80.26 & \spd{1.83} 
& 72.62 & \spd{1.27} 
& 68.00 & \spd{1.21} 
& \best{85.89} & \spd{1.78} 
& \best{88.30} & \spd{1.80} 
& \second{91.27} & \spd{1.65} 
& 80.53 & \spd{1.70} \\

\rowcolor{variantrow}
\quad {$\triangleright$}~SpecEyes (bottom-$r$) 
& 78.26 & \spd{2.18} 
& 80.26 & \spd{1.84} 
& \second{77.35} & \spd{1.05} 
& \second{72.31} & \slo{0.99} 
& \best{85.89} & \spd{1.81} 
& \best{88.30} & \spd{1.81} 
& \second{91.27} & \spd{1.73} 
& 81.95 & \spd{1.63} \\

\rowcolor{oursrow}
\quad {$\triangleright$}~\textbf{SpecEyes (min)} 
& \best{87.83} & \spd{1.32} 
& \best{82.89} & \spd{1.42} 
& \best{78.47} & \spd{1.01} 
& \best{73.31} & \slo{0.95} 
& 85.87 & \spd{1.77} 
& \best{88.30} & \spd{1.78} 
& \second{91.27} & \spd{1.70} 
& \best{83.99} & \spd{1.42} \\

\bottomrule
\end{tabular}}
\end{table*}

\subsection{Main Results}
\label{sec:main_result}

\cref{tab:main_results} compares \ours{} against the agentic baselines and 
SpecReason~\cite{pan2025specreason} across all seven evaluation splits, using 
two agentic backbones (DeepEyes~\cite{zheng2025deepeyes} and 
Thyme~\cite{zhang2025thyme}) paired with Qwen3-VL-2B~\cite{qwen3technicalreport} 
as the tool-free speculative model. For each \ours{} variant, we report the 
result at the best operating-point threshold that preserves the baseline level 
accuracy. Among the four confidence aggregation strategies, \ours{} (min) 
consistently delivers the strongest accuracy--speed profile, validating the 
worst-case guard design in \cref{sec:gating}; we focus the discussion on this variant below. With DeepEyes, \ours{} (min) achieves a 1.73$\times$ average speedup 
while \emph{improving} average accuracy from 81.39\% to 84.26\%. On 
V*~Bench~\cite{vstar}, it matches the baseline on Direct Attributes (90.43\%, 
1.53$\times$) and boosts the Relative Position subset from 82.89\% to 89.47\% at 
1.90$\times$. POPE benefits most (2.13--2.19$\times$) with accuracy consistently 
above baseline (\eg Adversarial: 78.43\%\,$\rightarrow$\,85.13\%), suggesting 
that bypassing unnecessary tool trajectories can also reduce hallucination errors. 
HR-Bench yields moderate speedups (1.08--1.13$\times$) because queries increasingly demand fine-grained, tool-assisted inspection.

Replacing the backbone with Thyme confirms generalization: \ours{} (min) 
yields a 1.42$\times$ average speedup while raising accuracy from 82.29\% to 
83.99\%. The per-benchmark pattern mirrors DeepEyes: POPE benefits most 
(1.70--1.78$\times$), V*~enjoys solid gains (1.32--1.42$\times$), and HR-Bench 
remains the bottleneck (0.95--1.01$\times$). The marginal sub-1$\times$ speedup 
on HR-Bench 8K arises because high-resolution inputs suppress both $\beta$ and 
$\alpha$, keeping $\beta\alpha$ low; in this regime the fixed cost of running 
$M_S$ slightly exceeds any savings, consistent with \cref{eq:e2elatency}.

In contrast, SpecReason~\cite{pan2025specreason} consistently \emph{decelerates} 
inference (0.37--0.61$\times$ with DeepEyes; 0.43--0.53$\times$ with Thyme), as 
the small model lacks structured tool-calling capability and incurs substantial 
token and turn overhead (414 tokens and 3.48 rounds on average). It also degrades 
sharply on POPE (as low as 49.10\%). By contrast, \ours{} lets accepted queries 
bypass the tool-use chain entirely, avoiding this overhead. The Qwen3-VL-2B 
(draft only) row establishes a speedup upper bound (4.13$\times$) at notable 
accuracy cost (78.93\%); \ours{} captures most of this latency saving while 
preserving full reasoning quality.

\begin{table*}[t]
\centering
\caption{Comparison of draft-only baselines and \textbf{SpecEyes (min)} on visual benchmarks. We report accuracy (\%) and speedup ($\times$) for two draft models ($M_s$). 
}
\label{tab:ablation_Ms_compact}
\setlength{\tabcolsep}{3.5pt}
\renewcommand{\arraystretch}{1.1}
\resizebox{\textwidth}{!}{%
\begin{tabular}{@{} ll cc cc cc c cc cc cc @{}}
\toprule
\multicolumn{2}{l}{\multirow{4}{*}{\textbf{Dataset}}} 
& \multicolumn{6}{c}{\textbf{$M_s$ = Qwen3-VL-8B}} & & \multicolumn{6}{c}{\textbf{$M_s$ = Qwen2.5-VL-7B}} \\
\cmidrule(lr){3-8} \cmidrule(lr){10-15}
& & \multicolumn{2}{c}{\multirow{2}{*}{Draft-only}} & \multicolumn{4}{c}{\textbf{SpecEyes (min)}} & & \multicolumn{2}{c}{\multirow{2}{*}{Draft-only}} & \multicolumn{4}{c}{\textbf{SpecEyes (min)}} \\
\cmidrule(lr){5-8} \cmidrule(lr){12-15}
& & \multicolumn{2}{c}{} & \multicolumn{2}{c}{DeepEyes} & \multicolumn{2}{c}{Thyme} & & \multicolumn{2}{c}{} & \multicolumn{2}{c}{DeepEyes} & \multicolumn{2}{c}{Thyme} \\
\cmidrule(lr){3-4} \cmidrule(lr){5-6} \cmidrule(lr){7-8} \cmidrule(lr){10-11} \cmidrule(lr){12-13} \cmidrule(lr){14-15}
& & Acc. & Spd. & Acc. & Spd. & Acc. & Spd. & & Acc. & Spd. & Acc. & Spd. & Acc. & Spd. \\
\midrule
\multirow{2}{*}{\textbf{V*}} 
& Attr. & 81.74 & \spd{4.23} & 92.17 & \spd{1.47} & 90.43 & \spd{1.28} & & 79.13 & \spd{3.95} & 90.43 & \slo{0.92} & 87.83 & \slo{0.93} \\
& Pos.  & 78.95 & \spd{1.72} & 80.26 & \spd{2.69} & 80.26 & \spd{1.57} & & 71.05 & \spd{1.68} & 78.95 & \spd{1.63} & 78.95 & \spd{1.25} \\
\midrule
\multirow{2}{*}{\textbf{HR-Bench}} 
& 4K    & 77.50 & \spd{2.46} & 78.49 & \spd{1.05} & 78.38 & \slo{0.96} & & 74.88 & \spd{2.75} & 75.97 & \spd{1.02} & 77.72 & \slo{0.85} \\
& 8K    & 69.90 & \spd{1.54} & 74.06 & \spd{1.06} & 74.22 & \slo{0.94} & & 66.87 & \spd{1.72} & 71.43 & \slo{0.98} & 72.31 & \slo{0.83} \\
\midrule
\multirow{3}{*}{\textbf{POPE}} 
& Adv.  & 84.47 & \spd{2.48} & 84.23 & \spd{1.75} & 85.52 & \spd{1.60} & & 79.66 & \spd{2.68} & 80.60 & \spd{1.20} & 82.82 & \spd{1.06} \\
& Pop.  & 86.67 & \spd{2.69} & 86.47 & \spd{1.80} & 87.27 & \spd{1.48} & & 81.23 & \spd{3.01} & 84.47 & \spd{1.27} & 86.20 & \spd{1.17} \\
& Rand. & 91.33 & \spd{3.06} & 89.70 & \spd{1.85} & 90.80 & \spd{1.56} & & 88.79 & \spd{3.55} & 89.77 & \spd{1.23} & 91.10 & \spd{1.19} \\
\midrule
\multicolumn{2}{l}{\textbf{Avg.}} & 81.51 & \spd{2.60} & 83.63 & \spd{1.67} & 83.84 & \spd{1.34} & & 77.37 & \spd{2.76} & 81.66 & \spd{1.18} & 82.42 & \spd{1.04} \\
\bottomrule
\end{tabular}}
\end{table*}

To isolate whether the gains come from \emph{routing} or merely from skipping 
tool calls, we evaluate the backbone ($\mathcal{M}_L$) with tools disabled 
(\textit{DeepEyes/Thyme w/o tools}).
Although this achieves high speedup (3.53$\times$ / 2.72$\times$ on average), 
accuracy collapses on POPE, which dropps from 78.43\% to 46.90\% on the Adversarial 
split with DeepEyes, because tool-required queries are forcibly answered without 
visual inspection.
\ours{} avoids this by routing only genuinely tool-free queries to $\mathcal{M}_S$ 
while preserving full agentic reasoning for the rest, achieving both strong 
accuracy and meaningful speedup simultaneously.

\subsection{Analysis of Confidence Calibration}
\label{sec:calibration}
\begin{figure}[t]
  \centering
  \includegraphics[width=\textwidth]{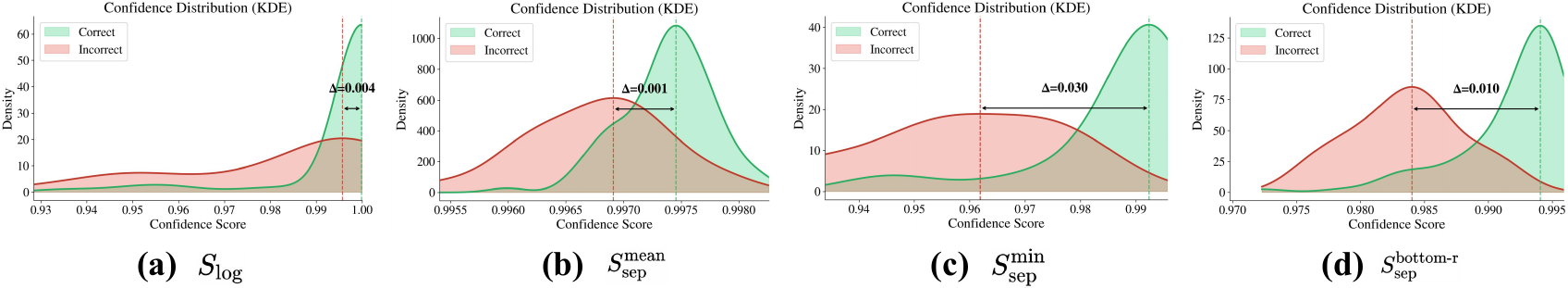}
\caption{\textbf{KDE of confidence scores for correct vs.\ incorrect samples on V*.} $\Delta$ measures gating discriminability via peak distance with Qwen3-VL-2B. Compared to the noticeable overlap in baselines (\textbf{a}, \textbf{b}, \textbf{d}), our \textbf{(c)}~$S_\text{sep}^\text{min}$ achieves the largest $\Delta$ with sharp bimodal separation, enabling an optimal accuracy-speed trade-off.}
  \label{fig:kde}
\end{figure}

A reliable gating signal must be \emph{discriminative}: confidence scores of 
correct answers should be stochastically higher than those of incorrect ones.
\cref{fig:kde} visualises this property via kernel density estimates (KDE) of 
each confidence score on correct and incorrect samples from $M_S$ on 
V*~\cite{vstar}, with each subplot annotated by $\Delta$ (peak distance between 
the two distributions) as a direct measure of discriminability.
Both $S_\text{log}$ (\cref{fig:kde}a) and $S_\text{sep}^\text{mean}$ 
(\cref{fig:kde}b) yield small $\Delta$: the former suffers from softmax 
overconfidence, and the latter is diluted by averaging over all tokens, leaving 
the two distributions heavily overlapping.
$S_\text{sep}^\text{bottom-r}$ (\cref{fig:kde}d) improves $\Delta$ by focusing on the lowest-separability tokens, yet residual overlap remains in the mid-range.
$S_\text{sep}^\text{min}$ (\cref{fig:kde}c) achieves the largest $\Delta$: incorrect samples collapse to a low-score peak, while correct samples form a sharp high-score mode, consistent with Proposition~1.
\cref{tab:main_results} shows that a single threshold to preserve accuracy while maximizing acceptance, explaining why \ours~(min) delivers a superior accuracy--speedup trade-off.

\subsection{Ablation Study}
\label{sec:ablation_study}

We study the effects of three key hyperparameters in \ours: the gating threshold, the serving batch size, and the separability computation parameter $K$.

\begin{figure}[t]
  \centering
    \includegraphics[width=\textwidth]{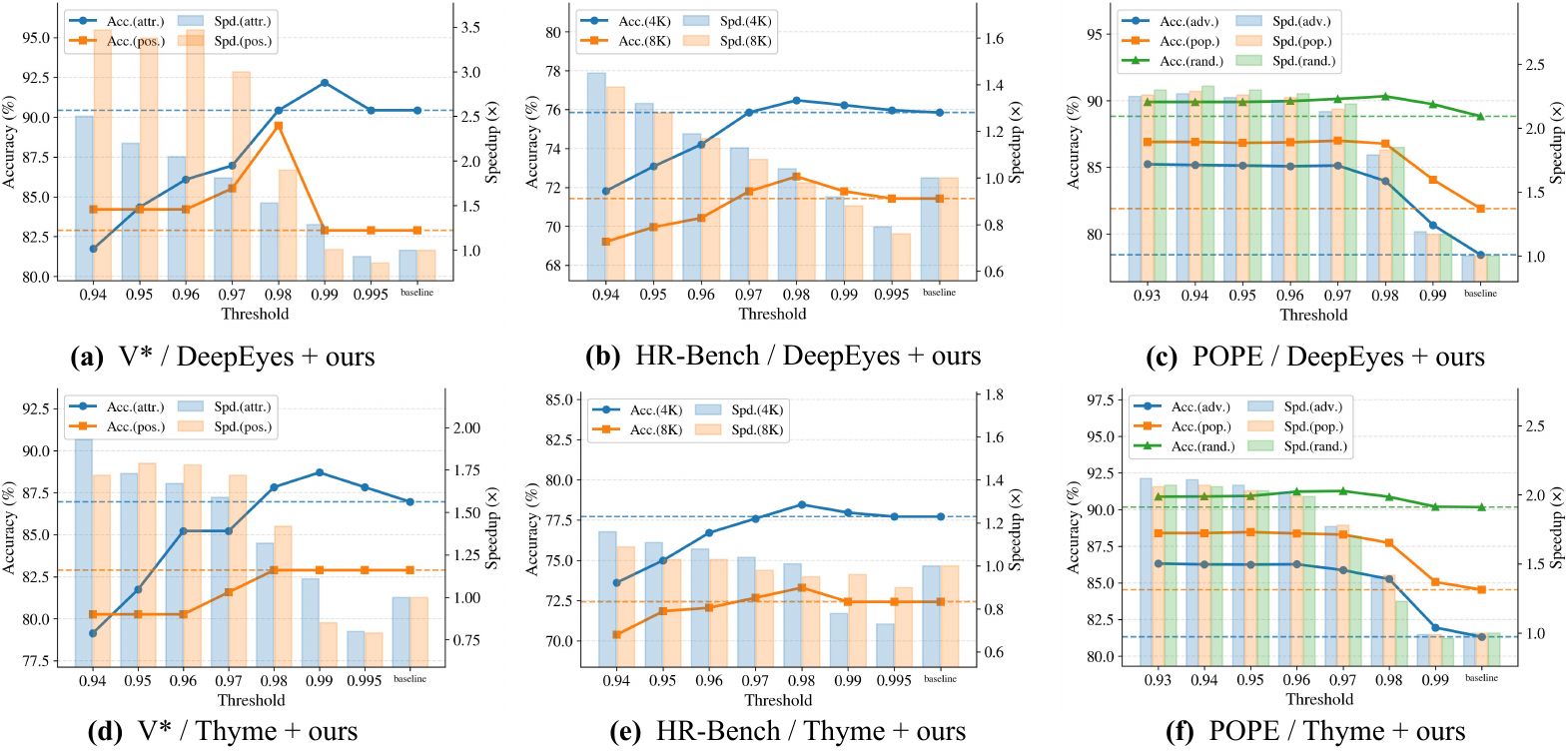}
    \caption{\textbf{Ablation on the gating threshold of SpecEyes.}
    Lowering the threshold increases speedup at cost of accuracy.
    Dashed horizontal lines indicate baseline accuracy.}
    \label{fig:ablation-thresholds}
    \vspace{-10pt}
\end{figure}

\myparagraph{Ablation on Gating Threshold.}
\cref{fig:ablation-thresholds} visualizes the accuracy--speedup trade-off as the 
gating threshold varies, using $S_\text{sep}^{\text{min}}$ across three benchmarks 
with both agentic backbones. Lower thresholds increase the acceptance ratio and 
speedup, while accuracy degrades gracefully. On V* and POPE, accuracy remains 
above or close to the agentic baseline over a wide range, indicating that many 
queries can be safely bypassed. HR-Bench is more sensitive, with limited speedup 
gains and accuracy drops below 0.97 due to a higher fraction of tool-required 
queries. Overall, \ours{} achieves a broad operating region that improves both 
accuracy and efficiency, showing that the threshold serves as a smooth control 
knob for navigating the accuracy--efficiency Pareto front.

\begin{figure}[t]
  \centering
    \includegraphics[width=\textwidth]{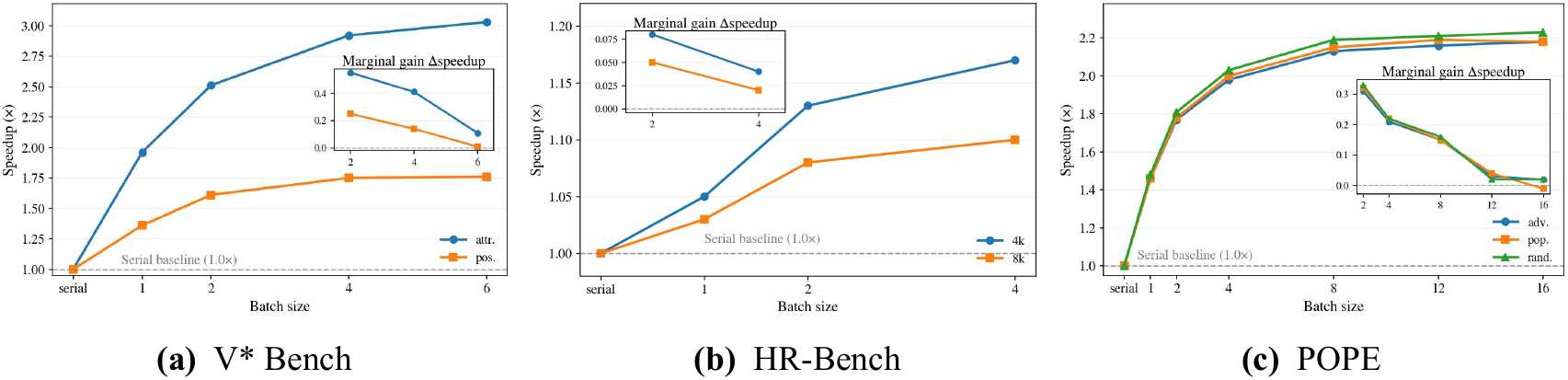}
    \caption{\textbf{Ablation on serving batch size.}
Larger batches amortize the stateless speculative stage, improving speedup with diminishing marginal gains as the stateful fallback bottlenecks. Curves report end-to-end speedup over the serial agentic baseline.}
  \label{fig:batchsize_ablation}
\end{figure}

\myparagraph{Ablation on Batch Size.}
\cref{fig:batchsize_ablation} studies the impact of serving batch size with the 
same gating threshold as the main results. Increasing batch size consistently 
improves end-to-end speedup without affecting accuracy, as batching only changes 
system execution. This follows from our heterogeneous funnel design 
(\cref{sec:parallel}): the stateless speculative stage is highly batchable, while 
the agentic fallback stage remains constrained by per-query tool dependencies, 
leading to diminishing gains at larger batches. Benchmarks with higher bypass 
rates benefit more from batching, whereas HR-Bench saturates 
earlier due to more tool-required queries.

\begin{figure}[t]
    \centering
    \includegraphics[width=\textwidth]{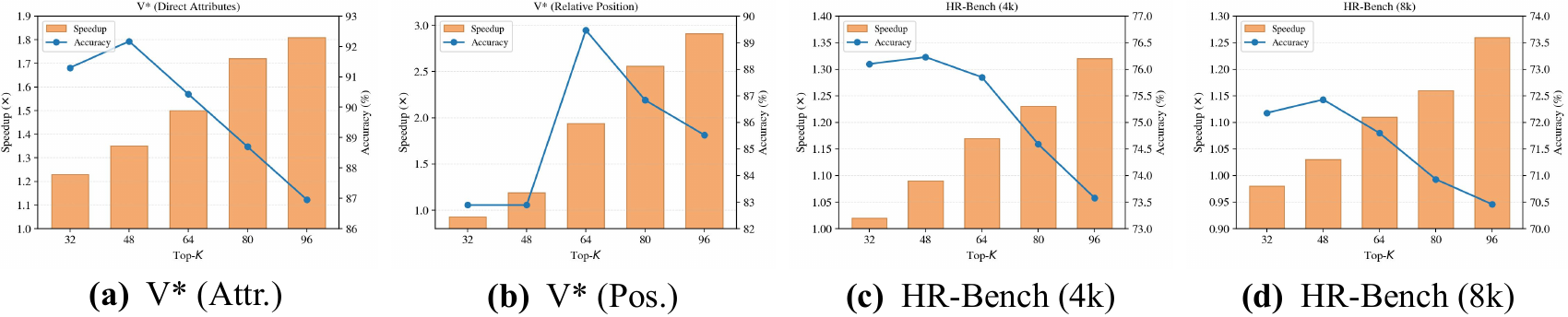}
    \caption{\textbf{Ablation on Top-$K$ in separability-based gating.} Increasing $K$ boosts speed but degrades accuracy, tuning the model's speculative aggressiveness.
    }
    \label{fig:topk_vstar}
    \vspace{-10pt}
\end{figure}

\myparagraph{Ablation on Top-$K$ in Separability Computation.}
As shown in \cref{fig:topk_vstar}, $K$ acts as a \emph{control knob}: increasing 
$K$ monotonically improves speedup but degrades accuracy, mirroring the effect of 
lowering the gating threshold, as larger $K$ includes tokens with weaker 
contrastive signal and thereby inflates confidence estimates.
We set $K{=}64$ as a balanced default, which matches baseline accuracy on Direct 
Attributes subsets and achieves a strong speedup on Relative 
Position (1.94$\times$), while overly large $K$ over-optimizes for speed 
at the cost of accuracy.

\myparagraph{Ablation on Draft Model.}
As shown in Tab.~\ref{tab:ablation_Ms_compact}, replacing Qwen3-VL-2B with a larger non-agentic model demonstrates that \ours{} 
is model-agnostic.
With Qwen3-VL-8B as $\mathcal{M}_S$, \ours{} achieves 
higher accuracy on V* (92.17\% Attr.) and HR-Bench 4K (78.49\%) under DeepEyes, 
showing that stronger speculators improve acceptance quality. However, the larger 
model reduces speedup to 1.67$\times$ (vs.\ 1.73$\times$ for 2B), making the 2B 
variant the better accuracy--efficiency trade-off. Qwen2.5-VL-7B follows a similar 
trend: despite comparable accuracy to the 2B variant, it achieves lower speedup 
(1.18$\times$ with DeepEyes and 1.04$\times$ with Thyme), indicating that 
$c_S$ becomes the main bottleneck for larger speculators.

\section{Conclusion and Future Work}
\label{sec:conclusion}

In this paper, we present \ours, an agentic-level speculative acceleration framework that lifts the speculation paradigm from individual tokens to the entire agentic pipeline. A lightweight, tool-free model speculatively answers queries that do not require multi-step tool use, governed by a \emph{cognitive gating} mechanism based on answer separability and served through a \emph{heterogeneous parallel funnel} that converts per-query latency savings into system-level throughput gains. Across three diverse image understanding benchmarks, \ours{} reduces end-to-end latency by up to $3.35\times$, while it is comparable with the agentic baseline in accuracy and delivers consistent throughput improvements under concurrent serving.
However, our speculative model currently operates at agentic depth $D{=}0$ (fully tool-free), limiting speedups on benchmarks (\eg HR-Bench) where most queries genuinely require tool assistance. A natural extension in future work is \emph{multi-depth speculation} ($D{=}1, 2, \ldots, n$), allowing the speculative model a bounded number of lightweight tool calls before gating, thereby intercepting queries at the earliest sufficient depth and further reducing unnecessary fallbacks.


\newpage

\bibliographystyle{splncs04}
\bibliography{main}
\end{document}